\documentclass[conference]{IEEEtran}
\IEEEoverridecommandlockouts
\usepackage{cite}
\usepackage{amsmath,amssymb,amsfonts}
\usepackage{graphicx}
\usepackage{textcomp}
\usepackage{xcolor}
\usepackage{marvosym}
\usepackage{multirow}
\usepackage[normalem]{ulem}
\useunder{\uline}{\ul}{}

\usepackage{etoolbox}
\makeatletter
\patchcmd{\@makecaption}
  {\scshape}
  {}
  {}
  {}
\makeatother

\usepackage{algorithm}
\usepackage{algorithmicx}
\usepackage{algpseudocode}

\usepackage{subfigure}
\usepackage{booktabs}

\def\BibTeX{{\rm B\kern-.05em{\sc i\kern-.025em b}\kern-.08em
    T\kern-.1667em\lower.7ex\hbox{E}\kern-.125emX}}
\begin{document}

\title{Digital Twin Mobility Profiling: A Spatio-Temporal Graph Learning Approach\\
}

\author{\IEEEauthorblockN{Xin Chen\IEEEauthorrefmark{1},
		Mingliang Hou\IEEEauthorrefmark{1}, Tao Tang\IEEEauthorrefmark{2}, Achhardeep Kaur\IEEEauthorrefmark{2} and
		Feng Xia\IEEEauthorrefmark{2}\textsuperscript{(\Letter)}}
	\IEEEauthorblockA{ \IEEEauthorrefmark{1}School of Software, Dalian University of Technology, Dalian 116620, China\\
		\IEEEauthorrefmark{2}School of Engineering, IT and Physical Sciences, Federation University Australia, Ballarat, VIC 3353, Australia\\
	\{xin.chen.jx; teemohold; tau.tang\}@outlook.com; achhardeepkaur@students.federation.edu.au; f.xia@ieee.org
}}


\maketitle

\begin{abstract}
    With the arrival of the big data era, mobility profiling has become a viable method of utilizing enormous amounts of mobility data to create an intelligent transportation system. Mobility profiling can extract potential patterns in urban traffic from mobility data and is critical for a variety of traffic-related applications. However, due to the high level of complexity and the huge amount of data, mobility profiling faces huge challenges. Digital Twin (DT) technology paves the way for cost-effective and performance-optimised management by digitally creating a virtual representation of the network to simulate its behaviour. In order to capture the complex spatio-temporal features in traffic scenario, we construct alignment diagrams to assist in completing the spatio-temporal correlation representation and design dilated alignment convolution network (DACN) to learn the fine-grained correlations, i.e., spatio-temporal interactions. We propose a digital twin mobility profiling (DTMP) framework to learn node profiles on a mobility network DT model. Extensive experiments have been conducted upon three real-world datasets. Experimental results demonstrate the effectiveness of DTMP.
\end{abstract}


\section{Introduction}

With the emergence of industrial revolution Industry 4.0, Industrial Internet-of-Things (IIoT) and smart city, Intelligent Transportation System (ITS) is anticipated to satisfy various technical demands and diverse, realistic requirements concerning accuracy, environment friendly and reliability~\cite{TIS2020,GreenTransportation}. Nowadays, ITSs are an essential topic from both a technological and a business standpoint. They mostly rely on Transportation Engineering models and algorithms, which are profoundly affecting human travel habits. Furthermore, the proliferation mobility data bring us to the era of transportation big data. To efficiently leverage mobility data, the concept of mobility profiling has been proposed as a promising way to design and virtualized ITS for urban traffic management and planning. Mobility profiling offers an effective simulation approach to meet the diverse use case requirements. It is a critical enabler for realising ITS application with  rigorous accuracy requirements on a cloud computing platform.

\par Mobility profiles are a summary of mobility-specific characteristics such as average speed and mobility flow~\cite{Amichi2019}. The process of extracting an entity's mobility patterns from data, such as a user's mobility interest and the crowd flow of a bus stop, is referred to as mobility profiling. Effective mobility profiling may aid our understanding of urban transportation, provide accurate predictions or projections, and increase the efficiency of government or policymakers' decision-making.
\par However, due to the high complexity of the industrial process of mobility data, directly employing mobility profiling in ITS suffers significant challenges. Existing ITS applications, in particular, are unable to capture the possible association retained in mobility data~\cite{Xia2018ICM}. It's essential to use mobility profiles to find urban mobility trends based on data acquired by IoT devices and to provide precise and reliable predictions. These predictions can be utilized to provide reliable recommendations to human planners and to carry out transportation management autonomously.

\begin{figure}[htbp]
    \centering
    \includegraphics[width=8cm]{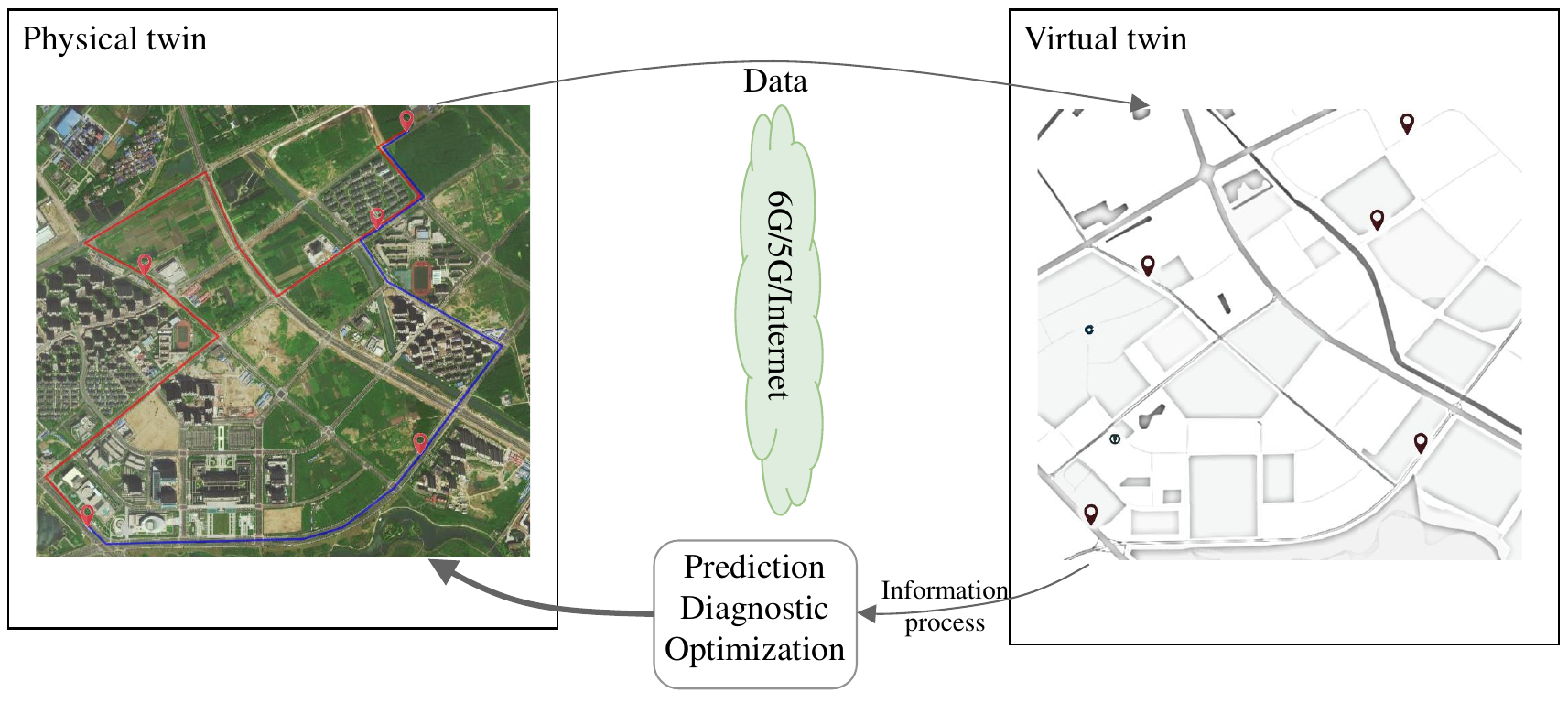}
    \caption{Virtual replica of physical transportation network using mobility network DT model.}
    \label{digital}
\end{figure}

\par The recent advancement in Digital Twin (DT) technology opens up a lot of possibilities for cyber-physical integration. Cyber-Physical System (CPS) is transformational technology that manage integrated systems between their physical assets and computing capability\cite{LEE201518}. DT is a technology that creates reflection of a physical net in the digital world. We agree that mobility profiling can greatly benefit from a mobility network DT. Firstly, a mobility DT creates a virtual replica of the physical transportation network, as shown in Fig. \ref{digital} and can be used to do a variety of data analyses on various what-if scenarios without disrupting the physical transportation network~\cite{BHATTI2021}. Secondly, a mobility network DT can generate accurate mobility profiles to support downstream prediction tasks after any configuration changes. A DT of the transportation network is crucial to achieve cost-efficient and performance-optimal transportation management, as well as to continue monitoring ITS performance under a set of operational circumstances without modifying the physical network.
\par
\begin{figure}[htbp]
    \centering
    \subfigure[Stations dependency under spatial and temporal.]{
    \includegraphics[width=8cm]{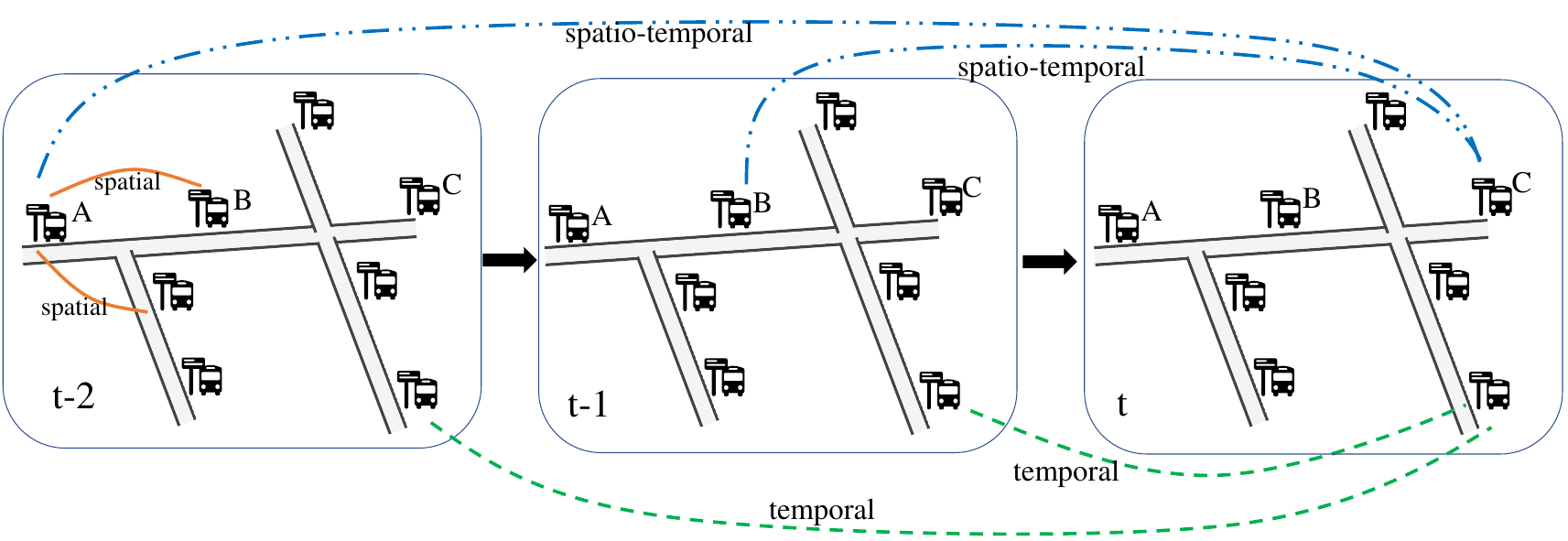}
    \label{map}
    }
    \subfigure[Flow from A, B and C stations. Stations A, B and C are on the same line and station C has some similarity in flow to stations A and B, with B having a smaller temporal shift and A having a larger temporal shift.]{
        \includegraphics[width=8cm]{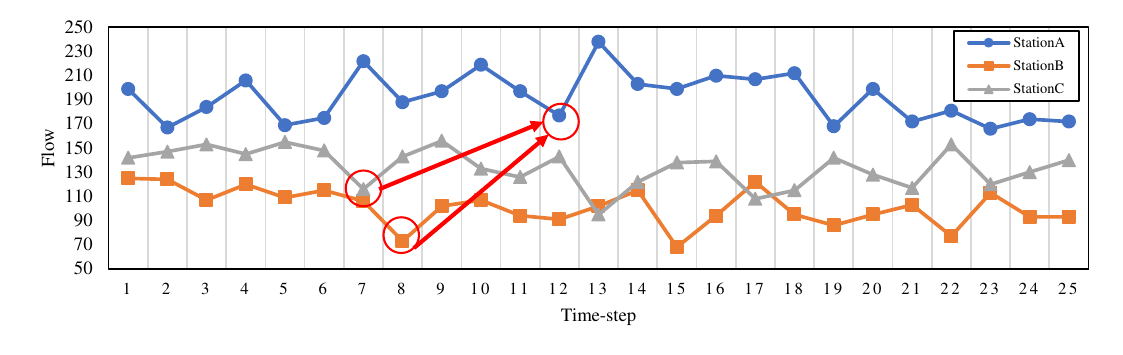}
    }
    \caption{An illustrative example of spatio-temporal dependency.}
    \label{introduction}
\end{figure}

 Artificial Intelligence (AI) techniques have recently advanced to the point where they can now meet the requirements of mobility network development~\cite{Xia2018ICM,Hou2020CSR}. When it comes to processing and analyzing large amounts of transportation data, cloud computing stands out as the most efficient method available~\cite{cloud2018,Xia2021TAI,Kong2021TII}. However, an intelligent DT system can only be developed by applying advanced AI techniques to the collected data. The Internet of Things (IoT) is used to collect massive mobility data from the physical world, and then use the data as input into an AI model to construct a digital twin. As a last phase, the developed DT can be used to optimize the performance of the ITS. On the other hand, transportation networks are fundamentally represented in graphs, many adaptive graph approaches~\cite{graphwavenet,AGCRN,wu2020connecting} have emerged to learn the spatio-temporal features based on Graph Neural Network (GNN). These approaches have achieved outstanding performance by introducing graph learning into their frameworks without a priory knowledge. In contrast to this, most existing approaches employ a graph to depict the interaction between nodes, which may not be suited to capture finely grained data-source specific patterns. As shown in Fig. \ref{introduction}, node relationships include sequence similarity relationships as well as temporal delay or lead relationships. Aside from that, current approaches~\cite{RNN1,DCRNN,ASTGCN} always seek to learn spatial and temporal dependencies individually, ignoring interactions between them. As a result, these models are limited in their ability to provide reliable profiling results on graph-structured data while capturing complicated spatio-temporal connections. This constraint limits the creativity of a mobility network DT and its implementation in ITS.

\par To address the aforementioned limitations, we design a mobility network profiling model (called DTMP) based on the Spatio-Temporal Graph Neural Network to discover the complicated and fine-grained spatial and temporal correlations in physical transportation networks and generate the accurate node profiles to support downstream analytical tasks in cloud platform. In particular, we develop a learning mechanism that is capable of learning complex mobility patterns without relying on domain knowledge of network structure. A supervised end-to-end learning process can be used to create the adaptive station profiles from the data collected. Furthermore, we design an unique alignment graph to help express fine-grained spatio-temporal interactions and propose a dilated alignment convolution network (DACN) to learn it. Finally, We employ gated Temporal Convolution Network (TCN)~\cite{graphwavenet} to assist the adaptive DACN to learn the temporal correlations.
\par In this paper, we present the spatio-temporal mobility network DT based on mobility data collected by automatic fare collection system deployed on buses. The mobility profiles refer to the learned virtual representations of bus stations, which can be utilized to support downstream analytical tasks. A summary of this paper's major contribution is given below:
\begin{enumerate}
    \item We propose to use spatial and alignment graphs to represent fine-grained spatio-temporal interactions and propose to alignment graph convolutional network (AGCN) to capture spatio-temporal dependencies in urban mobility.
    \item We propose a novel framework digital twin mobility profiling (DTMP) to learn node profiles on a mobility network DT. To develop accurate node profiles, the DTMP employs DACN and parallelizes gated TCN.
    \item We undertake comprehensive experiments on three real-world data sets. Experimental results demonstrate the effectiveness of the proposed framework. We have released the code on GitHub$\footnote{https://github.com/chenxino/DTMP}$.
\end{enumerate}
\par The remainder of the paper is set out as follows. We briefly cover related work in Section 2. The details of the proposed framework DTMP are given in Section 3. Performance evaluation will be presented in Section 4 where experimental results are analyzed. Finally, we conclude the paper in Section 5.

\section{Related Work}
This work is closely related to digital twins, mobility profiling and spatio-temporal graph learning. We briefly introduce previous work in this section.
\subsection{Digital Twins}
Digital twins can be defined as machines or computer-based models that are mirroring, emulating, simulating, or `twining' the life of a physical entity\cite{Bin2021, Steyn2021}. Specifically, a Digital Twin can be any of the following three types: 1) Digital Twin Prototype (DTP); 2) Digital Twin Instance (DTI); and 3) Digital Twin Aggregate (DTA)~\cite{DTsurvey}. A DTP is a completely digital model of a not been created thing, whereas a DTI is a virtual twin of an existing object that focuses on only one of its properties. A DTA is a collection of DTIs that may be an exact digital replica of a previously existing physical object.
\par Various types of digital twins are developed in ITS, including DT for automobile components, vehicles, vehicular networks, and road infrastructures. For example, Wang et al.~\cite{wangdt} proposed a framework for providing advisory speed assistance to the driver by vehicle-to-cloud (V2C) communication network DT. Cioroaica et al.~\cite{cior2019trust} developed a digital twin, called hardware-in-the-loop (vHiL) tested model to evaluate the trust-building capability of smart systems within an ecosystem. Chen et al.~\cite{chen2020UAV} proposed a framework to allocate complementary computation resources for mobile users in an mobile edge computational environment by utilizing unmanned aerial vehicles (UAVs). They developed a deep reinforcement learning (DRL) technique for task scheduling on the UAV and decreasing response time from the UAV to mobile users. The DRL model is trained by constructing a digital twin of the full mobile edge computational system. Kumar et al.~\cite{kumar2018novel} created an entire virtual model via digital twin to replicate the real-world scenario. Deep learning algorithms were used to predict driver behaviour.
\par As per best of our knowledge, we are the first to develop a station-based mobility network DT in order to extract precise bus station profiles to be used in downstream analytical operations. The DT we developed in this research can be classified as DTI because it only focuses on mobility-related aspects (spatio-temporal relations) of bus stations.
\subsection{Mobility Profiling}
Mobility profiling refers to the effort of extracting characteristics and features for a specific entity (user, station, area, etc.)\cite{BigTrajectoryData,Amichi2019}. Generally speaking, mobility profiling can be categorized into two classes: (1) static profiling and (2) dynamic profiling. Static profiling seeks to learn entity representations relying on varied mobility data that varies depending on the temporal perspective. For example, Ermal et al.~\cite{Pulse} proposed a subway station crowd flow prediction framework by integrating time-invariant station profiles such as remoteness from the downtown area, and mean flow, etc. Xia et al.~\cite{rankingstation} presented a framework to rank the station importance by considering the static subway profiles such as centrality, connectivity, etc. Dynamic profiling, on the other hand, relates to modeling the entity representations while taking into account the temporal elements that the profiles may alter over time. Zhao et al.~\cite{STELLAR} proposed a spatio-temporal latent ranking framework called STELLAR, capturing temporal impact on successive PoI (Point of Interest) recommendations. Wang et al.~\cite{mobilityprofiling} proposed an adversarial substructured representation learning method to extract mobility user profiles on mobile activity graphs. Wang et al.~\cite{mobilityprofiling2} proposed a framework by incorporating reinforcement learning into spatial knowledge graph representation learning to generate the dynamic mobility user profiles.

\subsection{Spatio-temporal Graph Learning}
The basic process of spatio-temporal learning is to model the spatial dependency in a set of spatial graphs with different time slots and then learn the temporal dependency of these spatial graphs by a temporal learning architecture, e.g., Recurrent Neural Networks (RNN)~\cite{RNN1}. Li et al.~\cite{DCRNN} designed the diffusion convolutional layer to extract the spatial features and adopted the Gated Recurrent Unit (GRU) to capture the temporal dependency. Ye et al.~\cite{Coupled_Graph} design the coupling mechanism to perform graph convolution in different layers using different graphs,  and then parasitize GCN to GRU to capture spatio-temporal dependencies. Yu et al.~\cite{STGCN} proposed stacked ST-Conv blocks for learning the spatio-temporal correlations on the predefined spatio-temporal graphs. The gated Convolutional Neural Networks (CNN) is utilized to learn the temporal dependency. However, predefined spatio-temporal graph learning could not capture the complex spatio-temporal patterns preserved in mobility data. Wu et al.~\cite{graphwavenet} proposed incorporating adaptive graph learning mechanism into fixed predefined graph learning to learn the spatio-temporal dependencies. Bai et al.~\cite{AGCRN} presented a full adaptive learning architecture, including node adaptive parameter learning and data adaptive graph generation, to learn the task-specific spatio-temporal features from mobility data. Li et al.~\cite{STFGNN} constructed a spatio-temporal fusion graph to model the spatio-temporal correlations explicitly. Then, it adopts gated dilated CNN to learn the local and global spatio-temporal correlations on the fusion graph. Oreshkin et al.~\cite{FC-GAGA} proposed a framework for learning spatio-temporal dependencies without using predefined graphs and relying only on learn able fully connected layers and gated graph architecture.

\par To learn station representations from mobility data, we use an adaptive graph learning technique, which is different from prior spatio-temporal graph learning research. Moreover, in order to learn the spatio-temporal interactions, we create and design a convolutional alignment graph network. Temporal dependency can be further learned with gated TCN in parallel with DACN.

\begin{table}[htp]
    \caption{Frequently used notations}
    \centering
    \renewcommand\arraystretch{1.2}
    \begin{tabular}{c|l}
        \toprule
        Symbol & Description \\
        \midrule
        $\mathcal{G}(V,A)$ & a traffic network \\
        $V=\{1,2, ...,N\}$ & set of nodes in the traffic network $\mathcal{G}$ \\
        $A \in \mathbb{R}^{N\times N}$ & predefined graph depends on prior knowledge\\
        $W_\theta = \{W, b, \Theta\}$ & set of learnable parameters in the model \\
        $X_\mathcal{G} \in \mathbb{R}^{T\times N\times C }$ & $T$-step historical feature matrix \\
        $Y_\mathcal{G} \in \mathbb{R}^{T'\times N\times C }$ & next $T'$-step traffic feature matrix \\
        $X^{(i)}_\mathcal{G} \in \mathbb{R}^{N\times C }$ & feature matrix in the $\mathcal{G}$ at time step $i$ \\
        $H \in \mathbb{R}^{T\times N\times F }$ & output of middle layer\\
        $A^{(S)} \in \mathbb{R}^{N\times N}$ & spatial adjacency matrix \\
        $A^{(T)} \in \mathbb{R}^{N\times N}$ & alignment adjacency matrix \\
        $A^{(S_i)} \in \mathbb{R}^{N\times N}$ & spatial adjacency matrix with $i$ shift \\
        $E_1, E_2 \in \mathbb{R}^{N\times e}$ & adaptive node embeddings\\
        \bottomrule
    \end{tabular}
    \label{notation}
\end{table}
\section{Design of DTMP}
\subsection{Spatio-temporal Prediction Problem Formulation}

\begin{figure*}[htbp]
    \centering
    \includegraphics[width=17cm]{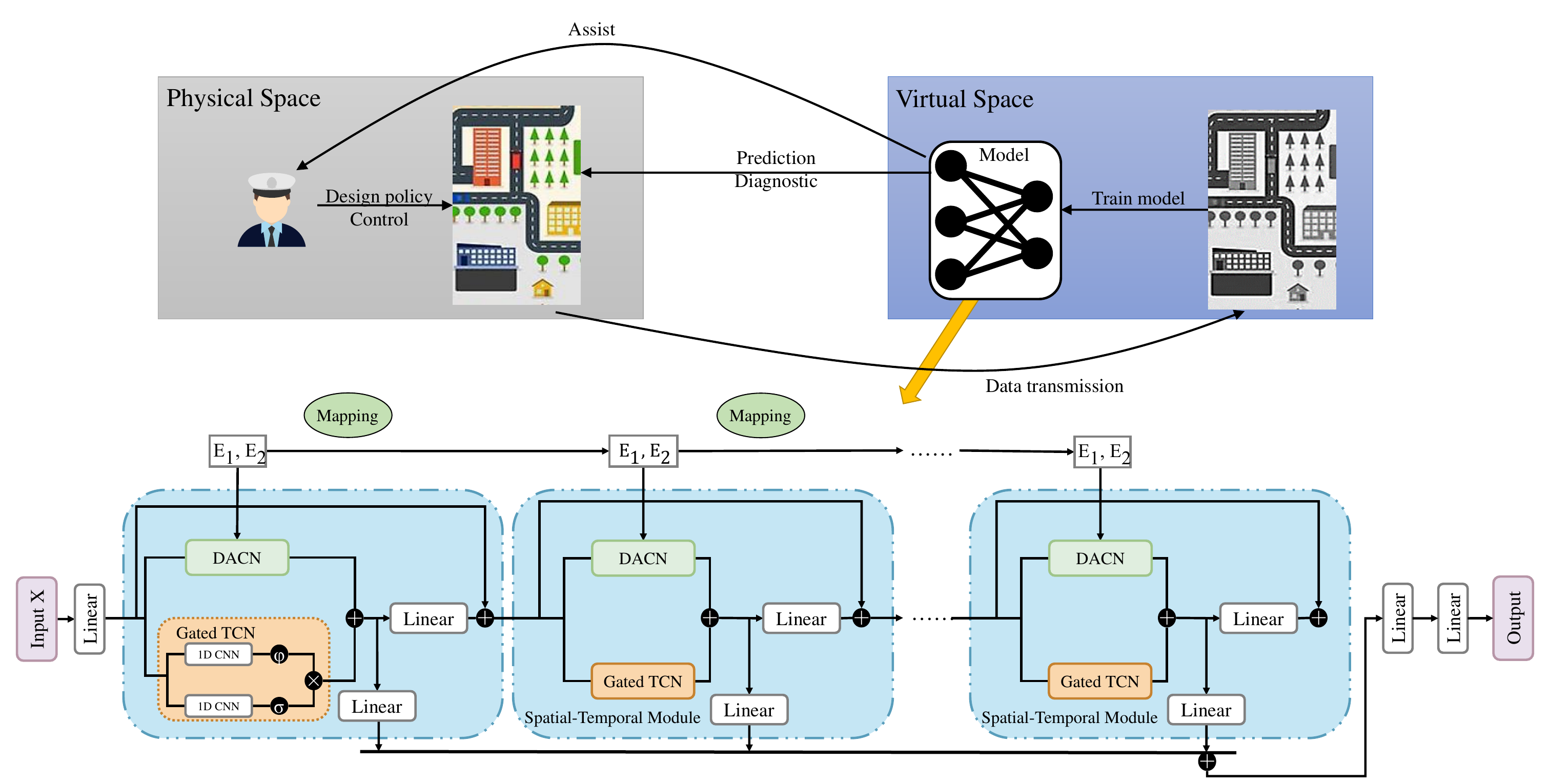}
    \caption{DTMP's framework. The model consists of multiple spatio-temporal modules and an output layer. Each spatial-temporal module transmits the result to the output layer through skip-connected.}
    \label{model}
\end{figure*}

In the transportation system, we define a graph $\mathcal{G}(V, A)$ to represent a traffic network, where $|V| = N$ is the set of nodes, $N$ denotes the number of vertices. $A\in \mathbb{R} ^{N\times N}$ is the matrix showing the proximity between the nodes. At time step $i$, the graph $\mathcal{G}$ has a feature matrix $X^{(i)}_{\mathcal{G}}\in \mathbb{R}^{N\times C}$, also called graph signals. $X^{(i)}_{\mathcal{G}}$ records the $C$ types of traffic features (such as flow, speed) of each nodes observed at time step $i$. We describe the traffic prediction problem as having $T$-step historical data $X_{\mathcal{G}}$, finding a mapping function $\mathcal{F} $ to predict the the next $T'$-step traffic condition $Y_{\mathcal{G}}$, formulated as:
\begin{equation}
    \begin{aligned}
        X_{\mathcal{G}} & = (X^{1}_{\mathcal{G}}, X^{2}_{\mathcal{G}}, ..., X^{(T)}_{\mathcal{G}})\\
        Y_{\mathcal{G}} & = (X^{(T+1)}_{\mathcal{G}}, X^{(T+2)}_{\mathcal{G}}, ..., X^{(T+T')}_{\mathcal{G}})\\
        Y_{\mathcal{G}} & = \mathcal{F}_{W_\theta}(X_{\mathcal{G}})\\
    \end{aligned}
    \label{Problem}
\end{equation}
where $W_\theta$ denotes all the learnable parameters in the model. Table \ref{notation} lists frequently used notations in this paper.

\subsection{Framework of DTMP}
We present the DTMP's framework in Fig. \ref{model}. The model is mainly composed of fully connected layers and stacked spatio-temporal modules. Each spatio-temporal module is mainly included of parallel DACN and gated TCN. Furthermore, coupling mapping transforms node embedding vectors between spatio-temporal modules, resulting in each module having its own graph structure. The node embedding is the station portrait learned by the model. By using skip-connected, each module transfers the result to the output layer. The spatio-temporal module's output is combined in the output layer, which translates it to the target dimension through two fully connected layers with activation functions. Spontaneous spatial and temporal information are extracted using DACN in the spatio-temporal module. Temporal dependencies are further captured using the gated TCN.

\subsection{Alignment Convolution Module}
The spatial correlation between nodes has always been regarded as the key factor and research hotspot in the traffic prediction problem. With the popularity of GNN, existing work mainly focuses on constructing an appropriate spatial graph and using GNN to capture the spatial dependencies in the traffic sequence\cite{Wang2020}. Relatively rare is the temporal relationship between the nodes taken into account. However, graph signals are time-series data. There should be a temporal relationship between two time series. We propose two types of graphs in the transportation network to better characterize the temporal and spatial relationships, spatial graph $A^{(S)}$ and alignment graph $A^{(T)}$. It's a spatial graph that describes the similarity of temporal patterns between nodes, similar to the prior work in this area as well, whereas, nodes in the alignment graph are sorted according to how far ahead or behind they are in the temporal pattern between them.

\paragraph{Alignment Graph}

\begin{figure}[htbp]
    \centering
    \includegraphics[width=7.5cm]{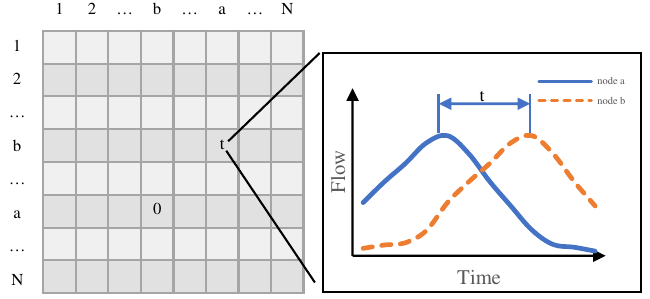}
    \caption{Alignment graph. Node b's traffic flow status is $t$ time steps behind node a.}
    \label{align}
\end{figure}

To design a time alignment graph, we use the temporal shift between graph signals. Nodes in the alignment graph can only be connected to their neighbors in the spatial graph since the alignment graph is dependent on the spatial graph. When two nodes are connected in a spatial network, we can conclude that they have similar graph signals, but this is only true after temporal lags or leads have been formed. Therefore, we build the alignment graph $A^{(T)}\in \mathbb{R} ^{N\times N}$ based on $A^{(S)}$ with each element $a^{(T)}_{i,j}\geqslant 0$, as shown in Fig. \ref{align}. There is an alignment value only for two points with edges in spatial graph. The $a^{(T)}_{i,j}$ represents the leading alignment of node $j$ relative to node $i$. The alignment graph is directed, if $a^{(T)}_{i,j}>0$ is true, then $a^{(T)}_{j,i}=0$.

\paragraph{AGCN}
We present a novel GNN titled alignment graph convolution network (AGCN), which implements graph convolution using the spatial graph and the alignment graph. General graph convolution does not employ alignment operation on the neighbor information, the spatial adjacency matrix to determine aggregated neighbors and aggregation pattern, whereas AGCN does. Here's how to calculate it:

\begin{equation}
    \operatorname{shift}(H, d) = H>>d
    \label{shift}
\end{equation}
\begin{equation}
    H^{(l+1)}_i = \sigma(\sum_{j=1}^n(\tilde{A}^{(S)}_{i,j} \times \operatorname{shift}(H^{(l)}_j, A^{(T)}_{i,j}))\times W  + b_i),
    \label{eq_AGCN}
\end{equation}
where $H^{(l+1)}_i$ is the output of node $i$ through AGCN, $\tilde{A}^{(S)}$ is the normalized spatial adjacency matrix, $W$ is learnable parameters and $b_i$ is bias. Shift operation is for rolling the graph signals along the time dimension. Elements that are shifted beyond the last position are discarded. As the formula \eqref{shift}, where $>>$ is right-shift operator and $d$ is the number of times the elements of $H$ have been shifted in time.

\par In traffic prediction, the convolution module can capture the temporal and spatial dependency concurrently, rather than learning them separately. There are, however, a couple of drawbacks to using AGCN on two graphs at the same time. Such graph convolution requires high computational costs because only one layer of the convolution network needs to complete $N\times N$ shift operations. The overhead caused by the shift operation makes the calculation speed extremely slow with limited computing resources. Another is the generation method of the graph. Through the distance between nodes, we can only get an inaccurate and incomplete alignment graph. It's not possible for a static, established network to include all the alignment information between nodes. Meanwhile, due to the discontinuous nature of the alignment graph, the adaptive generation method is likewise ineffective for producing the alignment graph during training.
\paragraph{DACN}
We propose an improved convolution network with a dilated aligned structure (DACN). First, we decompose the spatial graph according to the temporal alignment graph. Since the values of the alignment graph are discrete, the spatial graph is decomposed into multiple spatial graphs with the same shift value, which is expressed by the formula \eqref{graph_d}. Each sub-graph is initialized as an all-zero matrix, and then the spatial graph is assigned to the sub-graphs according to the alignment graph.
\begin{equation}
    A^{(S)} = \begin{cases}
        A^{(S_0)},&  \operatorname{If}\ A^{(T)}_{i,j}=0, A^{(S_0)}_{i,j}=A^{(S)}_{i,j} \\
        A^{(S_1)},&  \operatorname{If}\ A^{(T)}_{i,j}=1, A^{(S_1)}_{i,j}=A^{(S)}_{i,j} \\
        \vdots \\
        A^{(S_{(n-1)})},&  \operatorname{If}\ A^{(T)}_{i,j}=n-1, A^{(S_{(n-1)})}_{i,j}=A^{(S)}_{i,j},\\
        \end{cases}
    \label{graph_d}
\end{equation}

\begin{figure}[htbp]
    \centering
    \includegraphics[width=6cm]{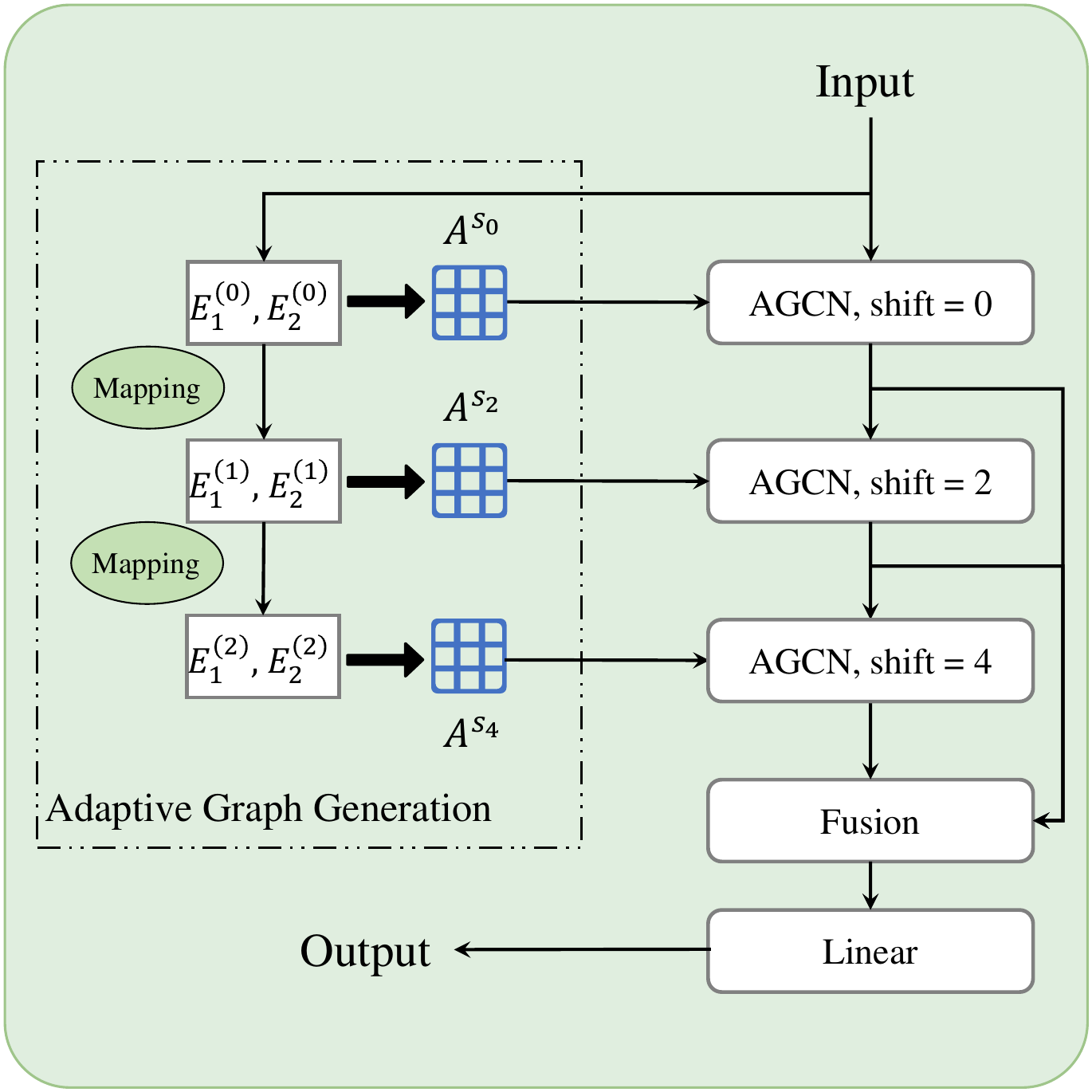}
    \caption{Dilated alignment convolution network and adaptive graph generation. With kernel size 3 and  dilation rate 2, it picks spatial graphs every 2 step and applies the AGCN. }
    \label{DACN}
\end{figure}

The original AGCN can be replaced by several AGCNs after the spatial graph has been decomposed. After decomposition, we construct a new DACN that uses spatial graphs produced from decomposition, as shown in Fig. \ref{DACN}. With a certain dilation rate $k$, we pick the spatial graph every $k$ step to apply AGCN to reduce the complexity of the model. Then, we connect AGCN in series according to the shift from small to large, and each AGCN is skip-connected to the output layer. After merging the outcomes of each AGCN(using concatenation-based fusion in the paper), the output layer converts each result to the relevant dimension. For each AGCN in DACN, since each spatial graph contains only one shift method, we can directly perform an overall shift on the input data and then conduct message propagation. So the formula \eqref{eq_AGCN} can be simplified as:
\begin{equation}
    H^{(l+1)}_i = \sigma(\tilde{A}^{(S_d)}_i\times\operatorname{shift}(H^{(l)}, d)\times W^{(l)}+b^{(l)}_i)
    \label{eq_DACN}
\end{equation}
where $\tilde{A}^{(S_d)}$ is the normalized spatial adjacency matrix with $d$ shift. $H^{(l)}$, $H^{(l+1)}$ represents the input and output of the $l$ layer respectively. $H^{(l+1)}_i$ is the output of node $i$.

\par Compared with the original alignment graph convolution, the number of shift operations performed in  DACN based on graph decomposition is reduced from the $\mathcal{O}(N^2)$ to $\mathcal{O}(1)$,  significantly reducing the computational overhead. On the other hand, time alignment graph information is allocated to each spatial graph, making it possible to obtain alignment graph information through learning. We overcome the problem of generating the discrete alignment graph. The graph generation problem has changed from constructing an alignment adjacency matrix and a spatial adjacency matrix to generating a series of spatial adjacency matrices with different alignment values.

\paragraph{Adaptive Graph Generation}
There are three main methods to generate the spatial graph: predefined graph by distance, predefined graph by time sequence similarity, adaptive generation graph by learn able parameters. The method of the predefined graph depends on prior knowledge, often unable to include all spatial information. Spatial correlation in the predefined graph has a deviation. Therefore, a new trend appears using an adaptive graph to describe spatial dependency information. For large data sets with many nodes, however, utilizing a direct adaptive method to generate numerous graphs can rapidly lead to an increase in parameters, which makes it more difficult to train the network and execute tests. So we use a coupled adaptive matrix generation method to generate a series of spatial adjacency matrices.
\par For zero-aligned spatial adjacency matrice, we randomly initialize two learnable node embedding vectors $\mathbf{E}^0_{1}, \mathbf{E}^0_{2}\in \mathbb{R}^{N\times e}$ for all node (for the dataset with prior knowledge, $\mathbf{E}^0_{1}$ and $\mathbf{E}^0_{2}$ can be obtained through the singular value decomposition of a given adjacency matrix), where $e$ is the embedding dimension of the node. The zero-aligned spatial graph can be calculated and normalized by the following formula:

\begin{equation}
    A^{(S_{0})} = \operatorname{ReLU}\left(\mathbf{E}_{1} \mathbf{E}_{2}^{(T)}\right)\\
\end{equation}
\begin{equation}
    \tilde{A}^{(S_{0})} = \operatorname{SoftMax}\left(A^{(S_{0})}\right)
\end{equation}

\par For the spatial graphs of other layers, we use the same generation method. Instead of using randomly generated node embedding, we use the previous layer of node embedding coupling mapping to obtain new node embedding. The coupled mapping formula is as follows:
\begin{equation}
    \begin{split}
        \mathbf{E}^{l+1}_{1} &= \mathbf{E}^l_{1}\times W +b\\
        \mathbf{E}^{l+1}_{2} &= \mathbf{E}^l_{2}\times W +b
    \end{split}
    \label{mapping}
\end{equation}
where $W\in \mathbb{R}^{e\times e} $ are learnable parameters and $b\in \mathbb{R}^{e}$ are bias. The coupling mapping of $\mathbf{E}_{1}, \mathbf{E}_{2}$ shapes parameters. There are $O(N\times N)$ parameters in the randomly generated graph, while the coupled graph only adds $O(2\times N\times e)$ parameters for each graph. In DACN, the adaptive graph generation method is shown in the right part of Fig. \ref{DACN}
\subsection{Gated Temporal Convolution Module}
According to the DACN module, a sequence of spatial graphs would be used to capture temporal and spatial dependencies, as well as similarity and shift information. However, adaptive learning is limited and the temporal dependence of the nodes themselves is also critical to traffic prediction. We employ the gated temporal convolution~\cite{graphwavenet} to extract the temporal dependency. We set DACN and gated TCN modules in the same layer to use the same dilation rate to make the gated temporal convolution module and alignment convolution module better cooperate. The gated temporal convolution module is defined as follows:
\begin{equation}
    H^{(l+1)}=\phi\left(\Theta_{1} \times H^{(l)}+b_1\right) \odot \sigma\left(\Theta_{2} \times  H^{(l)}+b_2\right)
    \label{eq_G_CNN}
\end{equation}
where $ H^{(l)}\in \mathbb{R}^{T\times N\times C}$ is the output of upper layer network, $\phi(\cdot )$, $\sigma(\cdot )$ are activation and sigmoid function respectively. $\Theta_{1}$, $\Theta_{2}$, $b_1$, and $b_2$ are model learnable parameters, $\odot$ is the Hadamard product.

\subsection{Loss Function}
We trained DTMP for traffic prediction using the Mean Absolute Error (MAE). Hence, the loss function of the model can be formulated as:
\begin{equation}
    L(W_\theta)=\frac{\sum_{i=1}^{T^{\prime}} |\mathbf{\widehat{Y}}_{\mathcal{G}}^{(i)}-\mathbf{Y}_{\mathcal{G}}^{(i)}| }{T^{\prime} \times N \times C},
    \label{loss}
\end{equation}
where $W_\theta$ denotes the set of learnable parameters in the model, $\mathbf{\widehat{Y}}_{\mathcal{G}}^{(i)}$ and $\mathbf{Y}_{\mathcal{G}}^{(i)}$ denote the predicted and true values in graph $\mathcal{G}$ at time $i$, respectively.
\par Algorithm \ref{A_model} illustrates the whole training procedure. $Linear(\cdot)$ represents a transformation on the feature dimension $C$ by fully connected layer.
\begin{algorithm}
	\caption{DTMP Training Process}
	\begin{algorithmic}[1]
		\Require The datasets D,  the number of spatio-temporal modules $L$, batch size is B;	
		\Ensure Well Trained DTMP;
		\State Initialize embedding vectors $\mathbf{E}_{1}, \mathbf{E}_{2}$.
        \Repeat
        \State sample a batch($X\in \mathbb{R}^{B\times T\times N\times C}$, $Y\in \mathbb{R}^{B\times T^{\prime}\times N\times C}$) from D.
        \State Expand the feature dimension of $X$ from $C$ to $F$ to get $H\in \mathbb{R}^{B\times T^{\prime}\times N\times F}$.
        \State Initialize temporary variable  $S$ to be null.
        \For{$i=1,2,\cdots , L$}
            \State Through formula \eqref{eq_DACN} and \eqref{eq_G_CNN}, calculate $Temp=DACN(H, E_1, E_2)+GatedTCN(H)$
            \State $H = H+Linear(Temp)$
            \State $S = S+Linear(Temp)$
            \State Convert $\mathbf{E}_{1}, \mathbf{E}_{2}$ through formula \eqref{mapping}
        \EndFor
        \State Obtain the prediction result $\widehat{Y}$ by executing two fully connected layers on $S$.
        \State Optimize the parameters by minimizing the loss \eqref{loss} between $Y$ and $\widehat{Y}$.
        \Until{convergence}

	\end{algorithmic}
	\label{A_model}
\end{algorithm}

\section{Experiments}

\subsection{Datasets}
We conduct experiments on three real-world traffic datasets to evaluate the performance of DTMP. PeMSD4 and PeMSD8 are collected from the Caltrans Performance Measurement System (PeMS). They refer to the San Francisco Bay Area and San Bernardino, California. After linear interpolation, the records are grouped into 5-minute windows. The Huaian dataset is generated in the City of Huaian, Jiangsu Province, China, by Panda Bus Company. We take the station as the node to extract the Spatio-temporal traffic data from bus transaction records and the corresponding bus arrival time records, then aggregated into 15-minute windows. Since the bus operating time is from 5 to 23 o'clock, a day contains 72-time steps in traffic flow. The detailed information is shown in Table \ref{datasets}.

\begin{table}[htp]
    \caption{Dataset description and statistics.}
    \centering
    \begin{tabular}{cccc}
        \toprule
        Data   & nodes & time steps & period                      \\
        \midrule
        PeMSD4 & 307   & 16692      & 1/Jan/2018 - 28/Feb/2018   \\
        PeMSD8 & 170   & 17856      & 1/Jul/2016 - 31/Aug/2016    \\
        Huaian & 721   & 2160       & 1/Apr/2020 - 30/Apr/2020 \\
        \bottomrule
    \end{tabular}
    \label{datasets}
\end{table}

\begin{figure*}[htp]
    \centering
    \subfigure[MAE]{
        \label{4MAE}
        \includegraphics[width=0.3\textwidth]{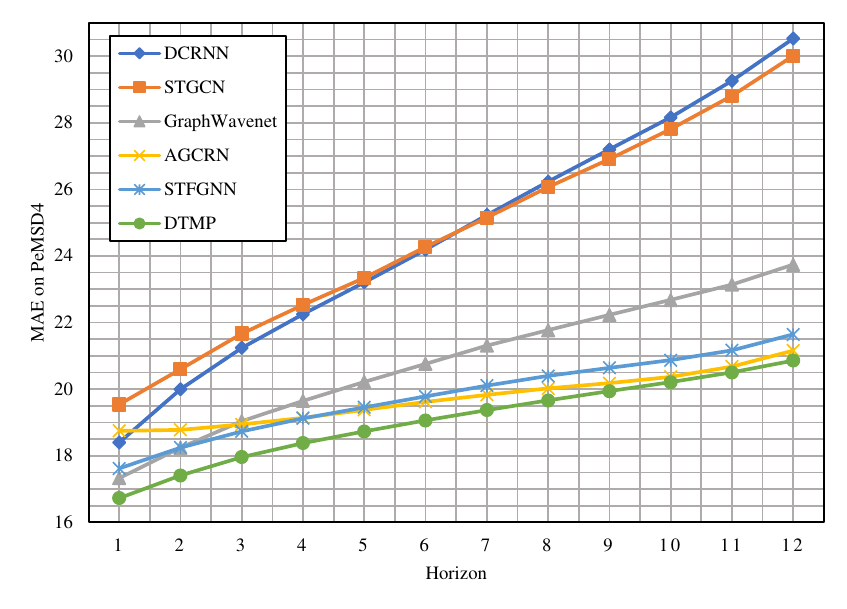}
    }
    \subfigure[MAPE]{
        \label{4MAPE}
        \includegraphics[width=0.3\textwidth]{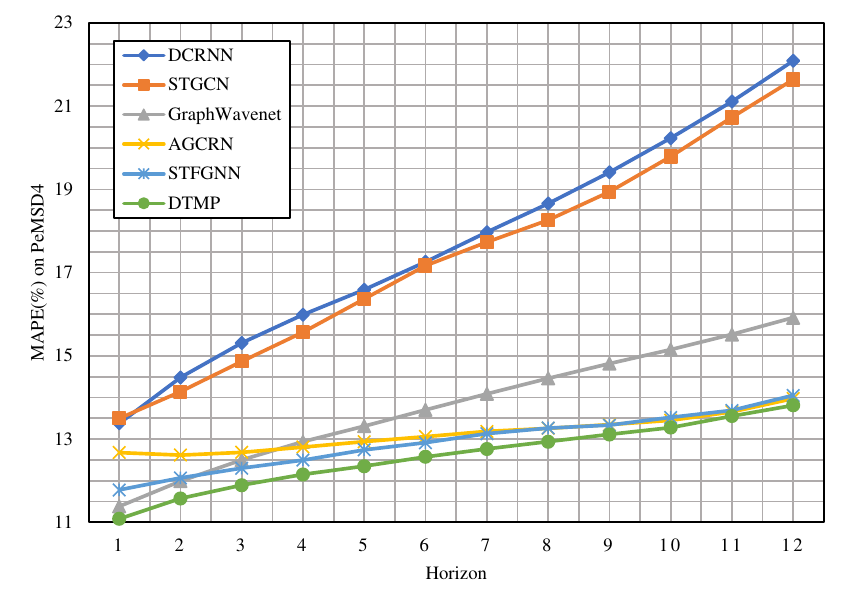}
    }
    \subfigure[RMSE]{
        \label{4RMSE}
        \includegraphics[width=0.3\textwidth]{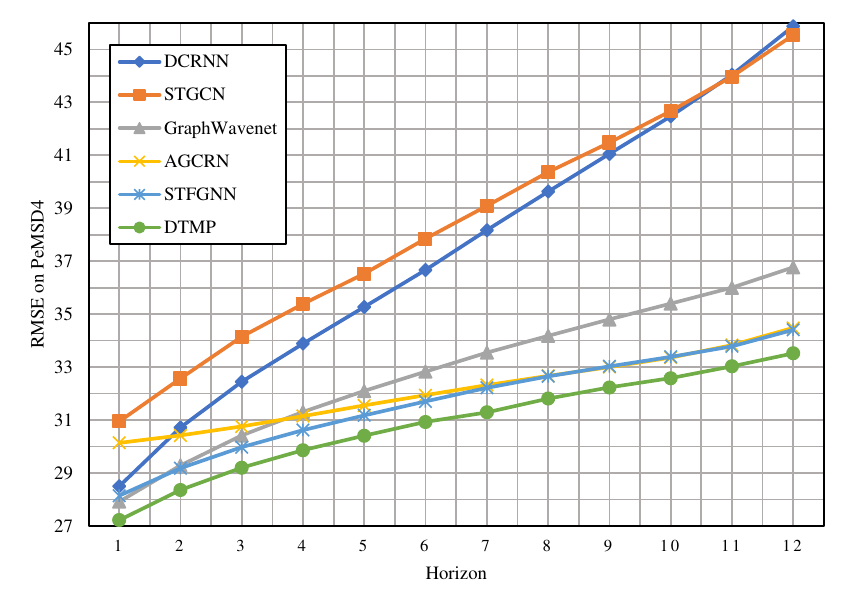}
    }
    \caption{Comparison of prediction performance for each horizon on the PeMSD4 dataset.}
    \label{4}
\end{figure*}

\subsection{Experimental Setup}
We divide the three datasets into training sets, validation sets, and test sets with a ratio of 6:2:2. The prediction problem chooses to use the 12-time steps historical data to predict the traffic conditions in the next 12- time steps. We implemented the code for DTMP traffic prediction in Python 3.7 with PyTorch 1.5.0. The model uses 6 layers of the spatio-temporal module, and the corresponding dilation rate is 1, 2, 4, 1, 2, 4, respectively. DACN's kernel size is set to 2, indicating that each DACN contains two general convolutions. Dropout with 0.3 is applied to the outputs of the DACN. Gated TCN's kernel size is set to 2. We initialize two sets of node embeddings randomly, and each embedding size is set to 10. In the training process, we optimize the model by Adam optimizer with a learning rate of 0.003. Mean Absolute Error (MAE), Root Mean Square Error (RMSE), and Mean Absolute Percentage Error (MAPE) are used to evaluate the performance of predictive models.

\subsection{Baselines}
We compare DTMP with the following models:
\begin{enumerate}
    \item HA: Historical Average model, which calculates the average of historical values to predict the future values.
    \item ARIMA \cite{ARIMA}: Auto-Regressive Integrated Moving Average model, which is a strategy for predicting time series. 
    \item GRU \cite{GRU}: Gate Recurrent Unit model, a sequence-to-sequence model that utilizes GRUs in encoder and decoder.  
    \item DCRNN (ICLR 2018) \cite{DCRNN}: Diffusion Convolution Recurrent Neural Network, which incorporates diffusion graph convolutions with GRU.
    \item STGCN (IJCAI 2018)\cite{STGCN}: Spatio-Temporal Graph Convolutional Network, which integrates graph convolutions with gated dilated convolutions.
    \item GraphWavenet (IJCAI 2019) \cite{graphwavenet}: Graph WaveNet, which uses adaptive adjacency to conduct graph convolutions and combines diffusion graph convolutions with gated dilated convolutions.
    \item AGCRN (NIPS 2020) \cite{AGCRN}: Adaptive Graph Convolutional Recurrent Network, which fusings node adaptive parameter learning and adaptive graph generation with GRU.
    \item STFGNN (AAAI 2021) \cite{STFGNN}: Spatial-Temporal Fusion Graph Neural Network, which combines fusion graph module of various spatial and temporal graphs with a novel gated convolution module.
\end{enumerate}

\subsection{Results and Analysis}
\subsubsection{Overall Comparison}
Table \ref{overall} shows the performance of DTMP and baseline models on PeMSD4, PeMSD8, and Huaian datasets. DTMP outperform in three datasets in terms of all evaluation metrics. It includes classic temporal methods including HA, ARIMA, and GRU with a large margin. Compared with other GCN-based spatio-temporal methods, DTMP surpasses predefined graphs (STGCN, DCRNN) and further improves the method of adaptive graphs (Graph WaveNet, AGCRN) with a significant margin. In PeMSD4 and PeMSD8 datasets, compared with AGCRN and STFGNN, our method achieves small promotion, but we obtain the bigger improvement in Huaian datasets. We consider the Huaian dataset has more nodes than the PeMSD datasets, and the nodes have more complex dependencies. DTMP has learned multiple adaptive graphs and has the ability to capture this complex relationship. In Fig. \ref{4}, we also display the evaluation measures at each horizon in the PeMSD4 dataset. Although AGCRN balances long-term and short-term predictions, our method's forecast errors are smaller than those on the 12 horizons.

\begin{table*}[htp]
    \centering
    \caption{Performance on PEMSD4, PEMSD8 and Huaian datasets.}
    \begin{tabular}{ccccccccccc}
        \toprule
        \multirow{2}{*}{model}  & dataset &  & PeMSD4 &  &  & PeMSD8 &  &  & Huaian &  \\ \cmidrule{2-11}
                                & metrics & MAE & RMSE & MAPE & MAE & RMSE & MAPE & MAE & RMSE & MAPE \\ \midrule
        \multicolumn{2}{c}{HA} & 38.30 & 56.76 & 39.63\% & 31.98  & 47.49 & 21.46\% & 127.4137 & 227.13 & 107.41\% \\ \midrule
        \multicolumn{2}{c}{ARIMA} & 26.10 & 34.40 & 41.70\% & 23.11 & 26.43 & 26.80\% & 82.54 & 94.69 & 68.85\% \\ \midrule
        \multicolumn{2}{c}{GRU} & 26.93 & 42.42 & 18.00\% & 20.72 & 32.28 & 15.05\% & 89.38 & 204.57 & 51.75\% \\ \midrule
        \multicolumn{2}{c}{DCRNN} & 24.66 & 37.76 & 17.71\% & 18.82 & 28.84 & 13.49\% & 80.22 & 148.44 & 52.03\% \\ \midrule
        \multicolumn{2}{c}{STGCN} & 24.80 & 38.69 & 17.42\% & 19.86 & 30.72 & 13.93\% & 72.96 & 134.92 & 43.50\% \\ \midrule
        \multicolumn{2}{c}{Graph Wavenet} & 20.40 & 32.41 & 13.83\% & 16.27 & 25.74 & 10.33\% & {\ul 56. 86} & {\ul 100.27} & {\ul 38.42\%} \\ \midrule
        \multicolumn{2}{c}{AGCRN} & 19.83 & 32.26 & 12.97\% & {\ul 15.95} & {\ul 25.22} & {\ul 10.09\%} & 70.92 & 168.63 & 41.72\% \\ \midrule
        \multicolumn{2}{c}{STFGNN} & {\ul 19.83} & {\ul 31.88} &  {\ul 13.02\%} & 16.64 & 26.22 & 10.06\% & 74.73 & 128.81 & 45.66\% \\ \midrule
        \multicolumn{2}{c}{\textbf{DTMP}} & \textbf{19.07} & \textbf{30.93} & \textbf{12.59\%} & \textbf{15.11} & \textbf{24.15} & \textbf{9.63\%} & \textbf{54.83} & \textbf{95.68} & \textbf{35.48\%} \\ \bottomrule
    \end{tabular}
    \label{overall}
\end{table*}

\subsubsection{Effect of Station Profiling}
\begin{figure}[htp]
    \centering
    \subfigure[Graph of station 0 for DACN.]{
    \label{relationship}
    \includegraphics[width=0.22\textwidth]{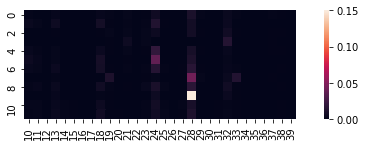}
    \includegraphics[width=0.22\textwidth]{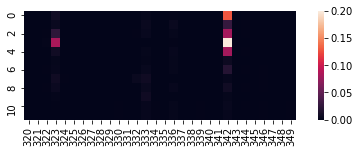}
    }

    \subfigure[Flow of stations 0, 28, 342]{
    \label{station_flow}
    \includegraphics[width=0.3\textwidth]{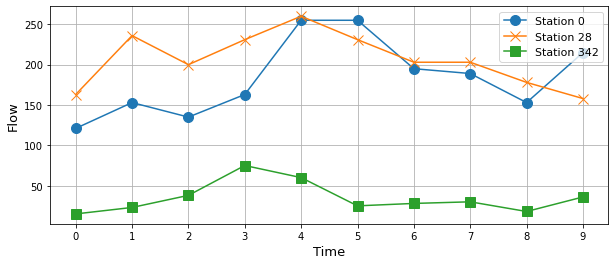}}
    \caption{The case of station 0. (a) The adaptively learned relationships between nodes, with 28 and 342 being the two most relevant station. (b) The traffic flow between these two stations and station 0.}
    \label{Fig.main}
\end{figure}

We use station 0 in the Huaian dataset as the research object and exhibit the relationships between it and other stations in 12 graphs to evaluate the efficacy of station profiling in DACN. These graphs are calculated by the station profiles (node embedding) and related model parameters after training. As shown in the Fig. \ref{relationship}, the $i$-th row represents the correlation strength with each station in the $i$-th graph. The X-axis represents station number. A stronger correlation with station 0 is found at stations 28 and 342. Fig. \ref{station_flow} shows the flow of stations 0, 28 and 342. We can see that compared with station 0, the flow at stations 28 and 342 have the same trend, but station 342 has a leading shift different from station 28. It demonstrates how station profiles capture the complex relationships between stations.

\subsubsection{Ablation Study}
We perform DTMP ablation experiments on PeMSD4 and Huaian datasets by eliminating many components. We design three variant models of DTMP as follows:
\begin{enumerate}
    \item No coupling mapping: DTMP without the coupling mapping between the spatio-temporal modules and between AGCNs. We utilize the same graph in different spatio-temporal modules.

    \item No alignment: DTMP without DACN module. We replaced the DACN with a ordinary GCN based on adaptive graph.
    \item No gated TCN: DTMP without gated TCN. We pass the outputs of DACN to fully connected layers ignore gated TCN.
\end{enumerate}
\par The results are shown in Table \ref{Ablation}, and we could conclude that the complete model possesses the best performance. Coupling mapping contributes to the model capture more complex Spatio-temporal relationships. DACN, compared with GCN, can capture temporal dependency. The introduction of DACN allows information to flow directly between different nodes at different times. It significantly improves performance on datasets with a large number of nodes. Gated TCN assists DACN to capture the temporal dependency further.

\begin{table}[htp]
    \centering  
    \caption{Ablation experiments on the PeMSD4 and Huaian datasets.}
    \begin{tabular}{ccccc}
    \toprule
    dataset                 & model elements      & MAE            & RMSE           & MAPE             \\ \midrule
    \multirow{4}{*}{PeMSD4} & No coupling mapping & 19.50          & 31.44          & 13.02\%          \\
                            & No alignment           & 19.95          & 32.00          & 13.31\%          \\
                            & No gated TCN             & 21.54          & 35.07          & 14.63\%          \\
                            & All                 & \textbf{19.07} & \textbf{30.93} & \textbf{12.59\%} \\ \midrule
    \multirow{4}{*}{Huaian} & No coupling mapping & 59.74          & 106.62         & 39.79\%          \\
                            & No alignment           & 61.77          & 114.30         & 38.97\%          \\
                            & No gated TCN             & 57.83          & 103.79         & 36.40\%          \\
                            & All                 & \textbf{54.83} & \textbf{95.68} & \textbf{35.48\%} \\ \bottomrule
    \end{tabular}
    \label{Ablation}
\end{table}

\subsubsection{Parameter Sensitivity}
To further investigate the model, we conduct parameter experiments on the PeMSD4 dataset, as shown in the Table \ref{Parameter}. In DACN, kernel size is one key parameter. It determines the number of adaptive graphs used in each DACN and the number of convolution operations performed. The table shows the model performance results when the kernel size is 1, 2, 3, and 4. The evaluation metrics first fall as the kernel size grows larger, indicating that expanding the kernel size can capture more full and intricate dependencies. When the kernel size exceeds 2, therewith, the metrics begin to rise. The reason for this is that the model has grown in complexity, making training more difficult. The node embedding records the dependencies between nodes and its dimension is another important parameter. The larger the dimension, the more information can be stored, but the more difficult it is to train the model. Therefore, the model performs best when the dimension is 10. In general, the model maintains a high performance under different parameters and shows the robustness of the model.

\begin{table}[htp]
    \centering
    \caption{Parameter experiments on the PeMSD4 dataset.}
    \begin{tabular}{ccccc}
    \toprule
    \multirow{2}{*}{Hyperparameter}     & dataset & \multicolumn{3}{c}{PeMSD4}                         \\ \cmidrule{2-5}
                                        & metrics & MAE            & RMSE           & MAPE
                                        \\ \midrule
    \multirow{3}{*}{Kernel size of DACN} & 1       & 19.43          & 31.29          & 12.94\%          \\
                                        & 2       & \textbf{19.07} & \textbf{30.93} & \textbf{12.59\%} \\
                                        & 3       & 20.01          & 32.02          & 13.15\%          \\ \midrule
    \multirow{5}{*}{Embedding dimension}             & 2       & 19.82          & 31.79          & 13.61\%          \\
                                        & 5       & 19.65          & 31.54          & 13.16\%          \\
                                        & 10      & \textbf{19.07} & \textbf{30.93} & \textbf{12.59\%} \\
                                        & 15      & 19.36          & 31.15          & 12.65\%          \\
                                        & 20      & 19.21          & 30.96          & 12.71\%          \\ \bottomrule
    \end{tabular}
    \label{Parameter}
\end{table}

\section{Conclusion}
For learning interactive spatio-temporal relationships, this paper has proposed a novel graph convolution network based on a spatio-temporal graph. Furthermore, we have designed an adaptive approach to generating alignment and spatial graphs based on mobility profiling and have proposed dilated alignment convolution network to simplify the calculation. It is combined with gated TCN and stacked to obtain the spatio-temporal graph learning framework DTMP for traffic forecasting and mobility profiling, and digital twin technology is used as a cost-effective tool. Extensive experiments and analysis have confirmed the model's improved forecasting performance compared to other techniques, as well as its ability to extract station information. An entirely new perspective on graph convolution networks was gained through the use of alignment procedures in order to discover time-dependent correlations.

\bibliographystyle{IEEEtran}
\bibliography{references}
\end{document}